\documentclass[10pt,twocolumn,letterpaper]{article}

\usepackage{cvpr}              
\definecolor{cvprblue}{rgb}{0.21,0.49,0.74}
\usepackage[pagebackref,breaklinks,colorlinks,allcolors=cvprblue]{hyperref}

\def\paperID{} 
\def\confName{CVPR}
\def\confYear{2026}

\usepackage{amsmath}
\usepackage{amsthm}
\usepackage{algorithm}
\usepackage{algorithmic}
\usepackage{bm}
\usepackage{threeparttable}
\usepackage{multirow}
\usepackage{makecell}
\usepackage{transparent}
\usepackage[dvipsnames]{xcolor}
\usepackage[table]{xcolor}

\newcommand{\faint}[1]{\transparent{0.65}\textcolor{gray}{#1}}

\theoremstyle{plain}
\newtheorem{theorem}{Theorem}[section]
\theoremstyle{definition}
\newtheorem{definition}[theorem]{Definition}

\title{Rethinking SNN Online Training and Deployment: Gradient-Coherent Learning via Hybrid-Driven LIF Model}

\author{Zecheng Hao\textsuperscript{1,2,3}\hspace{1.1em} Yifan Huang\textsuperscript{1,3}\hspace{1.1em} Zijie Xu\textsuperscript{1,3,4}\hspace{1.1em} Wenxuan Liu\textsuperscript{1,2}\thanks{Corresponding author.}\hspace{1.1em} Yuanhong Tang\textsuperscript{1}\\ Zhaofei Yu\textsuperscript{1,3,4}\hspace{1.1em} Tiejun Huang\textsuperscript{1,2,4}\\
\textsuperscript{1}School of Computer Science, Peking University\\
\textsuperscript{2}State Key Laboratory for Multimedia Information Processing, Peking University\\
\textsuperscript{3}Beijing Key Laboratory of Brain-inspired Spiking Large Models, Peking University\\
\textsuperscript{4}Institute for Artificial Intelligence, Peking University\\
{\tt\small\{haozecheng,yfhuang,liuwx66,ydtang,yuzf12,tjhuang\}@pku.edu.cn,zjxu25@stu.pku.edu.cn}
}

\begin{document}
\maketitle
\begin{abstract}
Spiking Neural Networks (SNNs) are considered to have enormous potential in the future development of Artificial Intelligence due to their brain-inspired and energy-efficient properties. Compared to vanilla Spatial-Temporal Back-propagation (STBP) training methods, online training can effectively avoid the risk of GPU memory explosion. However, current online learning frameworks cannot tackle the gradient discrepancy problem between the forward and backward process, merely aiming to optimize the GPU memory, resulting in no performance advantages compared to the STBP-based models in the inference stage. 
To address the aforementioned challenges, we propose Hybrid-Driven Leaky Integrate-and-Fire (HD-LIF) model family for efficient online learning, which respectively adopt different spiking calculation mechanism in the upper-region and lower-region of the firing threshold. We theoretically point out that our learning framework can effectively separate temporal gradients and address the misalignment problem of surrogate gradients, as well as achieving full-stage optimization towards learning precision, memory complexity and power consumption. Experimental results have demonstrated that our scheme is enable to achieve state-of-the-art performance for multiple evaluation metrics, breaking through the traditional paradigm of SNN online training and deployment. Code is available at \href{https://github.com/hzc1208/HD_LIF}{here}.
\end{abstract}   
\section{Introduction}
\begin{figure} [t]\centering  
\includegraphics[width=0.96\columnwidth]{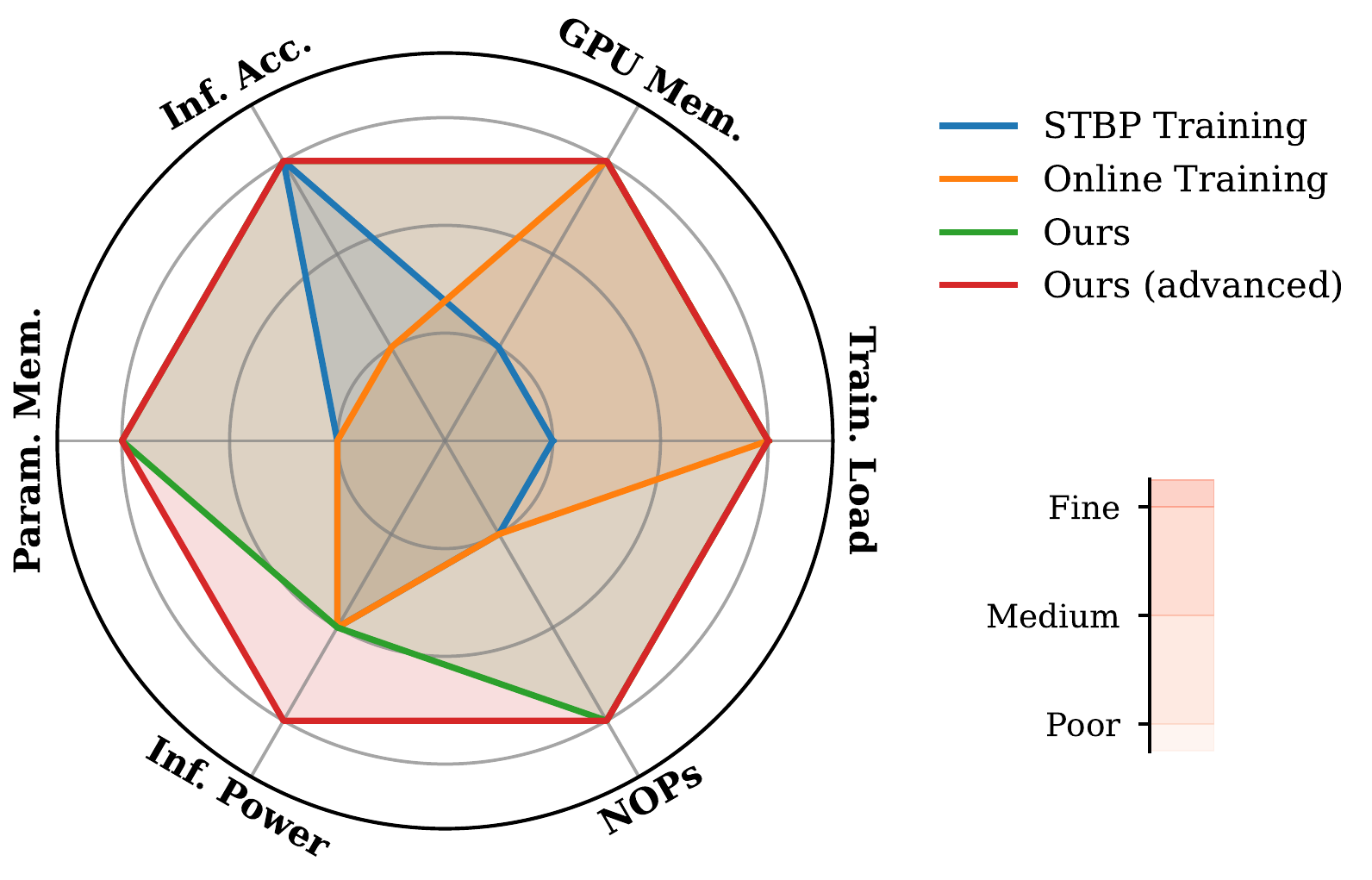}
\caption{In comparison of STBP and vanilla online learning paradigms, our schemes achieve comprehensive optimization towards multiple performance metrics of SNN training (Train. Load, GPU Mem.) and inference (Inf. Acc., Param. Mem., Inf. Power, NOPs) stages.}
\label{fig01}      
\end{figure}
Spiking Neural Networks (SNNs), as the third-generation neural network towards brain-inspired intelligence \cite{maas1997networks}, have gained widespread attention from researchers in Artificial Intelligence community. SNNs utilize spiking neurons as the basic computing unit to transmit discrete spike firing sequences to the post-synaptic layer. Due to the fact that spiking neurons only emit spikes when the membrane potential exceeds the firing threshold, compared to the activation values in traditional Artificial Neural Networks (ANNs), spike sequences have sparse and event-driven properties, which can demonstrate superior computational efficiency and robustness on neuromorphic hardware \cite{merolla2014million, davies2018loihi, pei2019towards, ding2025neuromorphic}.

Spatial-Temporal Back-propagation (STBP) is one of the most significant training algorithm in the supervised learning domain of SNNs currently \cite{wu2018STBP}. By introducing the concepts of temporal dimension and surrogate gradient, STBP simultaneously tackles the Markov property and non-differentiable problem of SNNs existed in the forward propagation and firing process. However, although STBP training has significantly improved the learning performance and generalization ability of SNNs \cite{wang2023ASGL, qiu2024gated, shi2024spikingresformer}, as its back-propagation chain has the property of temporal dependency, the GPU memory will inevitably boost linearly with the number of time-steps. This phenomenon greatly increases the training burden and hinders the further application of SNNs to complex scenarios \cite{kim2020spiking, yao2024spike} and advanced spiking models \cite{hao2023progressive, huang2024clif}.

To address this problem, researchers have transferred the idea of online learning to the STBP training framework \cite{xiao2022OTTT, Meng2023SLTT}, which means that by detaching the temporal dependent gradient terms, SNNs can immediately perform back-propagation at any time-step. This scheme ensures that the corresponding GPU memory is independent of the training time-steps and remains constant, effectively alleviating the problem of computation memory explosion. However, as shown in Fig.\ref{fig01}, the current proposed methods based on online learning still have the following defects: ({\romannumeral1}) the degradation of inference accuracy caused by the discrepancy between forward and backward propagation, ({\romannumeral2}) inference deployment without advantages.

The reason for the first defect is that the surrogate function of spiking neurons is generally related to the value of membrane potential and the spike sequence is usually unevenly distributed in the temporal dimension, making the temporal dependent gradients different from each other and unable to merge with the back-propagation chain along the spatial dimension. In this case, when online learning frameworks detach temporal dependent gradients, it will lead to inconsistency between forward and backward propagation, while the gradient misalignment inherent in the surrogate function will further intensify this phenomenon, resulting in learning performance degradation. The second defect refers to the fact that current online learning methods merely focus on optimizing training GPU memory. However, the obtained SNN does not have additional advantages on various performance metrics compared to the corresponding model trained by STBP in the inference stage, which seriously damages the practical application value of online learning.

Based on the above discussion, we propose the concept of Hybrid-Driven Leaky Integrate-and-Fire (HD-LIF) model family for efficient online learning. On the one hand, HD-LIF model can effectively separate backward gradients from the temporal direction and integrate them into the spatial dimension; on the other hand, it can be efficiently combined with various techniques for optimizing SNN inference overhead, achieving plug-and-play online training and deployment. Our contributions are summarized as follow:
\begin{itemize}
    \item Compared to vanilla spiking models, we point out that HD-LIF model has more superior gradient separability and alignment from the theoretical perspective, which is conducive to achieving high-performance online training.
    \item The model family covers multiple variant models adapted to different scenarios. Among them, Parallel HD-LIF can significantly reduce Neural Operations (NOPs) in the inference stage; Mem-BN HD-LIF, which is the version based on membrane potential batch-normalization, can effectively regulate the degree of gradient separability and be re-parameterized into vanilla HD-LIF model, as well as combining with efficient attention mechanism.
    \item For various datasets with different data-scale and data-type, our learning framework achieves state-of-the-art (SoTA) performance in multiple dimensions. For example, we reach top-1 accuracy of 78.61\% on CIFAR-100 dataset under the condition of saving $10\times$ parameter memory, $11\times$ inference power and 30\% NOPs.
\end{itemize}
\section{Related Works}
\begin{figure*} [t]\centering  
\includegraphics[width=1.8\columnwidth]{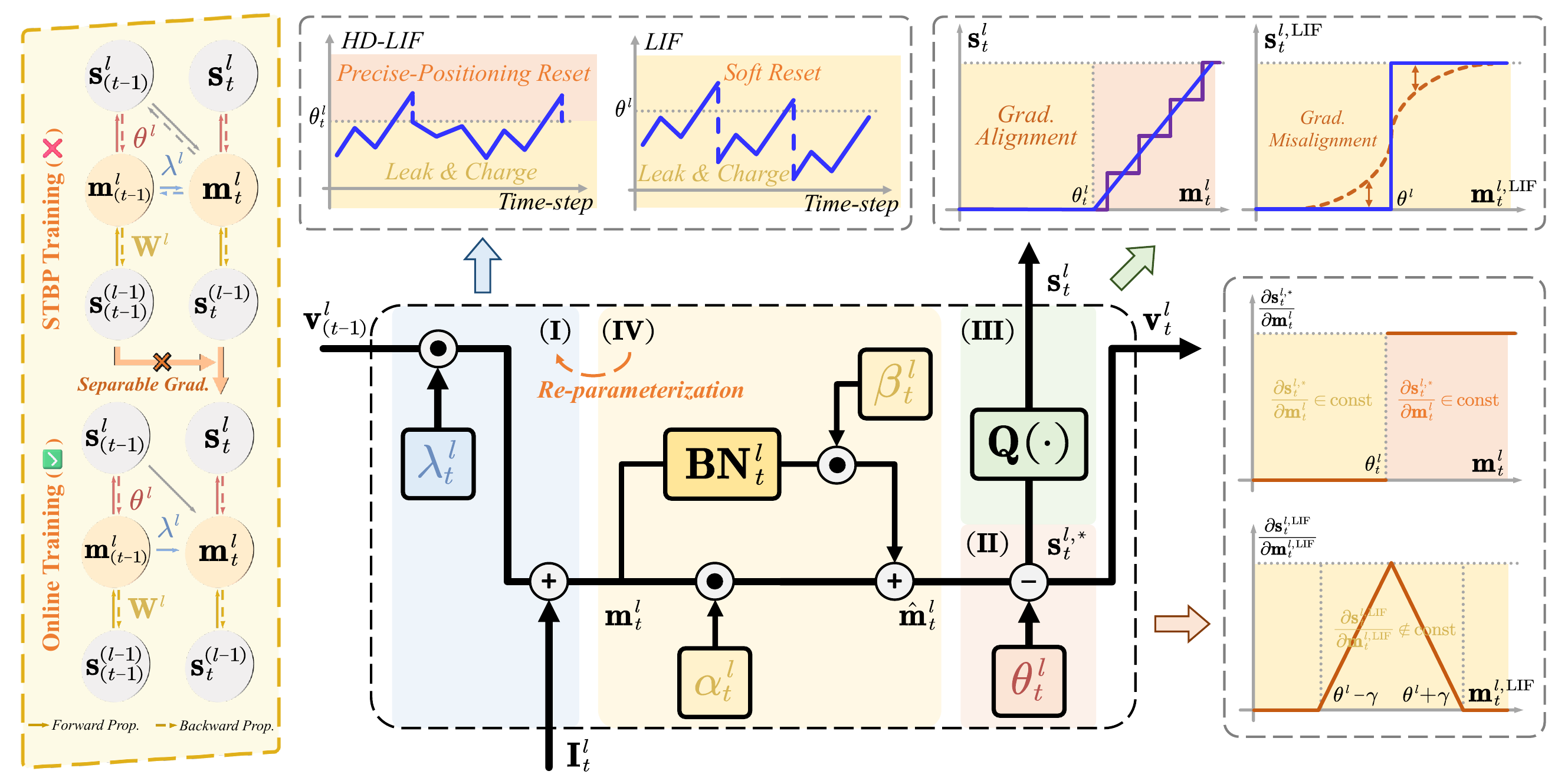} 
\caption{An overview of HD-LIF model. Compared to vanilla LIF, HD-LIF (middle-bottom) can enhance the gradient separability (right-bottom) and alignment (right-top) through the hybrid-driven spiking calculation mechanism (middle-top), thereby overcoming the discrepancy existed between the forward and backward process of conventional online training (left).}
\label{fig02}      
\end{figure*}
\textbf{Recurrent learning algorithms for SNNs.} Considering the similarity in computational mechanisms between SNNs and Recurrent Neural Networks (RNNs), Wu \etal \cite{wu2018STBP} and Neftci \etal \cite{neftci2019surrogate} transferred the Back-propagation Through Time (BPTT) method from RNNs to the supervised learning field of SNNs and utilized surrogate functions to tackle the non-differentiable problem existed in the spike firing process, which is called the STBP training algorithm. On this basis, Li \etal \cite{li2021dspike}, Guo \etal \cite{guo2022recdis} and Wang \etal \cite{wang2023ASGL} respectively attempted to start from the perspective of regulating the distribution about the backward gradient and membrane potential, introducing progressive surrogate functions and penalty terms. Tang \etal \cite{tang2022relaxation} proposed Relaxation LIF to mitigate gradient mismatch in deep SNN training. Deng \etal \cite{deng2022temporal} proposed a target learning function which comprehensively considers the SNN output distribution within each time-step, which is particularly suitable for neuromorphic sequential data. To further improve the learning stability and performance of SNNs, various BatchNorm (BN) modules \cite{zheng2021going, duan2022TEBN, Guo2023MBPN, jiangtab2024} and attention mechanisms \cite{yao2023attention, qiu2024gated, zhou2023spikformer, yao2024spike} have been proposed successively, which capture the representation information contained in spike sequences from multiple dimensions, including spatial-wise, temporal-wise and channel-wise. Recently, advanced spiking models have become a focus of academic attention. Researchers have proposed a variety of neuron models with stronger dynamic properties and memorizing capabilities around membrane-related parameters \cite{fang2020incorporating, yao2022GLIF}, firing mechanism \cite{guo2024ternary, shen2024conventional} and dendrite structure \cite{hao2023progressive, wang2024autaptic}, promoting deeper exploration towards brain-inspired intelligence. 
LIAF-Net \cite{wu2021liaf} adopt a similar model structure to this work and combine it with a series of RNN-related modules, but they aim to achieve lightweight information processing rather than efficient online training and deployment.
In addition, a spatial-temporal back-propagation algorithm based on spike firing time \cite{bohte2002error, kim2020unifying, zhang2020temporal, zhu2023exploring} has also attracted widespread attention. However, this series of methods are currently limited by high computational complexity and unstable training process, which cannot be effectively applied to complex network backbones and learning scenarios.

\textbf{Online learning algorithm for SNNs.} Although STBP learning algorithm promotes SNNs to join the club of high-performance models, it also brings severe computational burden to SNNs during the training phase, especially the GPU memory that will increase linearly with the number of time-steps. Xiao \etal \cite{xiao2022OTTT} transferred the idea of online learning to the domain of SNN direct training, which splits the back-propagation chain by ignoring the backward gradients with temporal dependencies, making the training GPU memory independent of time-steps. On this basis, Meng \etal \cite{Meng2023SLTT} proposed a selective back-propagation scheme based on online learning, which significantly improves training efficiency. Yang \etal \cite{yang2022tandem} combined online learning with ANN-SNN knowledge distillation, further accelerating the training convergence speed of SNNs. Zhu \etal \cite{zhu2024online} proposed a brand-new BN module suitable for online learning, which enhances the stability of gradient calculation by considering the global mean and standard deviation in the temporal dimension. To enrich the neurodynamic property of online learning, Jiang \etal \cite{jiang2024ndot} introduced the difference of membrane potential between adjacent time-steps as a feature term into the backward gradient calculation. Inspired by the architecture of reversed network, Zhang \etal \cite{zhang2024memory} and Hu \etal \cite{hu2024high} respectively proposed reversible memory-efficient training algorithms from spatial and temporal perspectives. This type of algorithm can ensure computational consistency between online and STBP learning under the condition of occupying constant GPU memory, but it requires bi-directional computation towards all intermediate variables, which inevitably increases computational overhead. Alternatively, FPTT with liquid time-constant neurons \cite{yin2023accurate} enables accurate online training of long sequences with constant memory.
\section{Preliminaries}
\textbf{Leaky Integrate-and-Fire (LIF) model.} The current mainstream spiking model used in SNN community is LIF model \citep{gerstner2002spiking}, which involves three core calculation processes, \ie charging, firing and resetting. As shown in Eq.(\ref{p-eq01}), at each time-step, LIF model will receive the input current $\mathbf{I}_t^{l,\text{LIF}}$ and refer to the previous residual potential $\mathbf{v}_{(t-1)}^{l,\text{LIF}}$, then accumulate the corresponding membrane potential $\mathbf{m}_t^{l, \text{LIF}}$. When $\mathbf{m}_t^{l, \text{LIF}}$ has exceeded the firing threshold ${\theta}^l$, a spike $\mathbf{s}_t^{l,\text{LIF}}$ will be transmitted to the post-synaptic layer and $\mathbf{m}_t^{l, \text{LIF}}$ will be reset. Here $\mathbf{W}^l$ denotes the synaptic weight and $\lambda^l$ represents the membrane leakage parameter.
\begin{align}
    \mathbf{m}_t^{l, \text{LIF}} &= \lambda^l \odot \mathbf{v}_{(t-1)}^{l,\text{LIF}} + \mathbf{I}_t^{l,\text{LIF}}, \
    \mathbf{v}_t^{l,\text{LIF}} = \mathbf{m}_t^{l, \text{LIF}} - \mathbf{s}_t^{l,\text{LIF}}, \nonumber \\
    \mathbf{I}_t^{l,\text{LIF}} &= \mathbf{W}^l\mathbf{s}_t^{(l-1),\text{LIF}}, \
    \mathbf{s}_t^{l,\text{LIF}} = \left\{
        \begin{aligned}
        &\theta^l,\ \mathbf{m}_t^{l, \text{LIF}} \geq {\theta}^l \\
        &0,\ \text{otherwise}
        \end{aligned}
    \right..
    \label{p-eq01}
\end{align}
\textbf{STBP Training.} To effectively train LIF model, the back-propagation procedure of SNNs usually chooses to expand along both spatial and temporal dimensions. We use $\mathcal{L}$ to denote the target loss function. As shown in Eq.(\ref{p-eq02}), $\frac{\partial \mathcal{L}}{\partial \mathbf{m}_t^{l, \text{LIF}}}$ will depend on both $\frac{\partial \mathcal{L}}{\partial \mathbf{s}_t^{l,\text{LIF}}}$ and $\frac{\partial \mathcal{L}}{\partial \mathbf{m}_{(t+1)}^{l,\text{LIF}}}$ simultaneously, while the non-differentiable problem of $\frac{\partial \mathbf{s}_t^{l,\text{LIF}}}{\partial \mathbf{m}_t^{l, \text{LIF}}}$ is tackled through calculating approximate surrogate functions.
Although STBP training enables SNN to achieve relatively superior performance, it inevitably causes severe GPU memory overhead during the training process, which will increase linearly with the number of time-steps.
\begin{align}
    \frac{\partial \mathcal{L}}{\partial \mathbf{m}_t^{l, \text{LIF}}} &= \underbrace{\frac{\partial \mathcal{L}}{\partial \mathbf{s}_t^{l,\text{LIF}}}\!\frac{\partial \mathbf{s}_t^{l,\text{LIF}}}{\partial \mathbf{m}_t^{l, \text{LIF}}}}_{\textit{spatial dimension}} + \underbrace{\frac{\partial \mathcal{L}}{\partial \mathbf{m}_{(t+1)}^{l,\text{LIF}}}\!\frac{\partial \mathbf{m}_{(t+1)}^{l,\text{LIF}}}{\partial \mathbf{m}_t^{l, \text{LIF}}}}_{\textit{temporal dimension}}.\\
    \nabla_{\mathbf{W}^l} \mathcal{L} &= \sum_{t=1}^T \frac{\partial \mathcal{L}}{\partial \mathbf{m}_t^{l, \text{LIF}}} \frac{\partial \mathbf{m}_t^{l, \text{LIF}}}{\partial \mathbf{W}^l}, \nonumber \\
    \frac{\partial \mathbf{m}_{(t+1)}^{l,\text{LIF}}}{\partial \mathbf{m}_t^{l, \text{LIF}}} &= \lambda^l + \frac{\partial \mathbf{m}_{(t+1)}^{l,\text{LIF}}}{\partial \mathbf{s}_t^{l,\text{LIF}}} \frac{\partial \mathbf{s}_t^{l,\text{LIF}}}{\partial \mathbf{m}_t^{l, \text{LIF}}}.
    \label{p-eq02}
\end{align}
\textbf{Online Training.} To avoid the issue of training memory overhead, a feasible solution is to detach $\frac{\partial \mathcal{L}}{\partial \mathbf{m}_{(t+1)}^{l,\text{LIF}}}\frac{\partial \mathbf{m}_{(t+1)}^{l,\text{LIF}}}{\partial \mathbf{m}_t^{l, \text{LIF}}}$ from the gradient calculation graph, thereby making the back-propagation chain independent in the temporal dimension. Online training enables SNNs to update gradients at any time-step, keeping the GPU memory at a constant level.
\section{Methods}
\begin{figure*} [t]\centering  
\includegraphics[width=1.8\columnwidth]{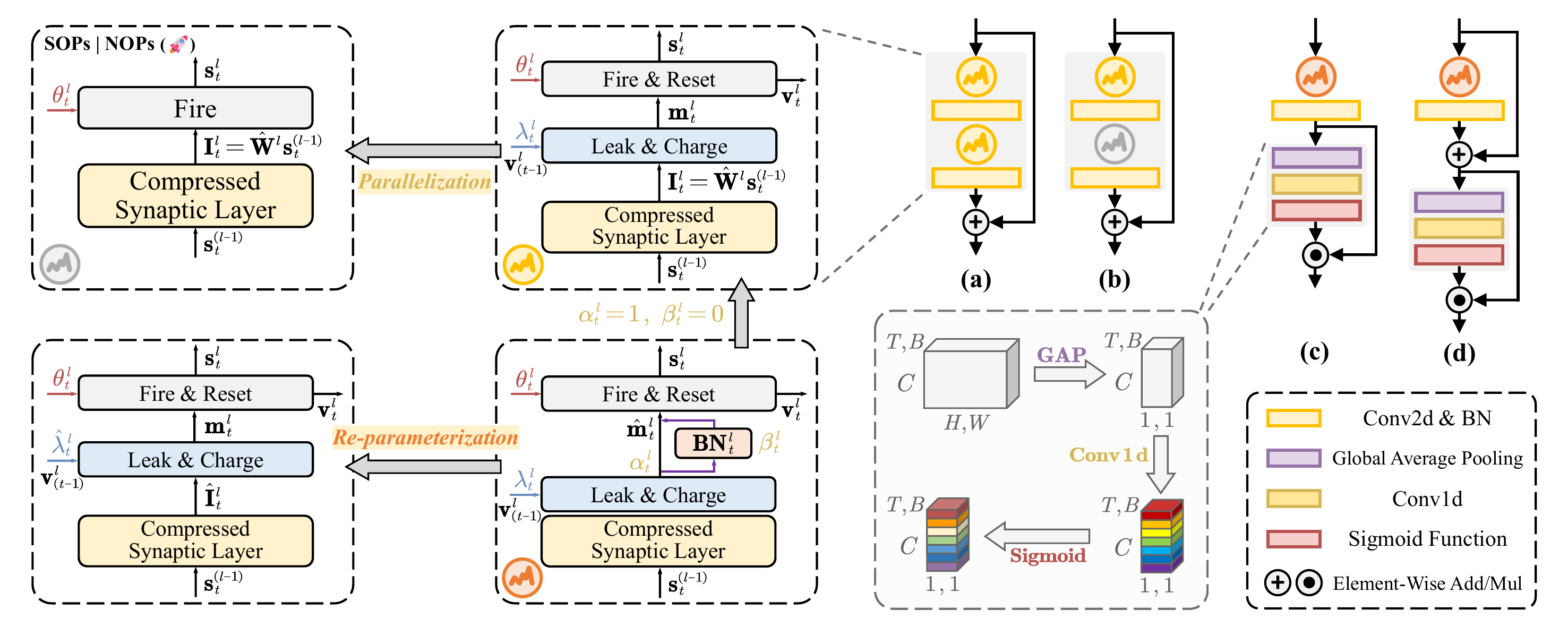} 
\caption{The structural description for SNN blocks based on HD-LIF model family. (a): vanilla version, (b): parallel version (w/ Parallel HD-LIF), (c)-(d): $\text{SECA}_\text{\uppercase\expandafter{\romannumeral1}}$ and $\text{SECA}_\text{\uppercase\expandafter{\romannumeral2}}$ (w/ Mem-BN HD-LIF).}
\label{fig03}
\end{figure*}
\subsection{Overcoming the back-propagation discrepancy of online training}
\label{section01}
Eq.(\ref{p-eq02}) unfolds the STBP back-propagation chain recursively from back to front along the temporal dimension, but one can find that the backward gradients can also be rewritten as a cumulative form: $\frac{\partial \mathcal{L}}{\partial \mathbf{m}_t^{l,\text{LIF}}} = \frac{\partial \mathcal{L}}{\partial \mathbf{s}_t^{l,\text{LIF}}} \frac{\partial \mathbf{s}_t^{l,\text{LIF}}}{\partial \mathbf{m}_t^{l,\text{LIF}}} + \sum_{i=t+1}^T \frac{\partial \mathcal{L}}{\partial \mathbf{s}_i^{l,\text{LIF}}} \frac{\partial \mathbf{s}_i^{l,\text{LIF}}}{\partial \mathbf{m}_i^{l,\text{LIF}}} \prod_{j=t+1}^i \frac{\partial \mathbf{m}_j^{l,\text{LIF}}}{\partial \mathbf{m}_{(j-1)}^{l,\text{LIF}}}$. On this basis, we first point out the concept of Separable Backward Gradient:
\begin{definition}
\label{def01}
Assume $\frac{\partial \mathcal{L}}{\partial \mathbf{s}_{1}^{l,\text{LIF}}} = ... = \frac{\partial \mathcal{L}}{\partial \mathbf{s}_{T}^{l,\text{LIF}}}$, we can rewrite $\frac{\partial \mathcal{L}}{\partial \mathbf{m}_t^{l,\text{LIF}}} = \frac{\partial \mathcal{L}}{\partial \mathbf{s}_{t}^{l,\text{LIF}}} \sum_{i=t}^T \bm{\epsilon}^l[i,t]$, here $\bm{\epsilon}^l[i,t]$ denotes the temporal gradient contribution weight of the $i$-th step \wrt the $t$-th step:\\
$\bm{\epsilon}^l[i,t] \in \left\{
    \frac{\partial \mathbf{s}_i^{l,\text{LIF}}}{\partial \mathbf{m}_i^{l,\text{LIF}}} \prod_{j=t+1}^i \frac{\partial \mathbf{m}_j^{l,\text{LIF}}}{\partial \mathbf{m}_{(j-1)}^{l,\text{LIF}}},\ i>t ;\
    \frac{\partial \mathbf{s}_t^{l,\text{LIF}}}{\partial \mathbf{m}_t^{l,\text{LIF}}},\ i=t
    \right\}.$
\\
If $\forall t, i\in [1,T]$, $\bm{\epsilon}^l[i,t]$ is a constant value, then we can further have $\left( \frac{\partial \mathcal{L}}{\partial \mathbf{m}_t^{l,\text{LIF}}} \right)_{\textit{Online}} \Leftrightarrow \left( \frac{\partial \mathcal{L}}{\partial \mathbf{m}_t^{l,\text{LIF}}} \right)_{\textit{STBP}}$. Here $\Leftrightarrow$ denotes the equivalence of gradient calculation. The gradient at this point is called Separable Backward Gradient.
\end{definition}
When the precondition of Definition \ref{def01} holds true, the back-propagation chain can be considered separable in the temporal dimension and the backward gradient of online training can be seamlessly transformed from that of STBP training. Unfortunately, as shown in Fig.\ref{fig02}, vanilla STBP training generally requires surrogate gradient functions which are related to the membrane potential value (\eg Triangle Function: $\frac{\partial \mathbf{s}_{t}^{l,\text{LIF}}}{\partial \mathbf{m}_t^{l,\text{LIF}}} = \frac{1}{\gamma^2}\max\left(\gamma - |\mathbf{m}_t^{l,\text{LIF}} - \theta^l|, 0\right)$), causing $\bm{\epsilon}^l[i,t] = \mathcal{F}(\mathbf{m}_t^{l,\text{LIF}},...,\mathbf{m}_i^{l,\text{LIF}})$. Considering that the specific values of membrane potential along temporal dimension are unpredictable and irregular, $\bm{\epsilon}^l[i,t]$ will not satisfy the condition of constant value. Therefore, current online training cannot effectively overcome the discrepancy between forward and backward propagation, which limits its learning precision.

To tackle this problem, we first propose vanilla HD-LIF, which is an advanced spiking model suitable for online training. In Eq.(\ref{eq01}) and Fig.\ref{fig02}, compared to vanilla LIF model, HD-LIF model retains the inherent membrane potential calculation mechanism of spiking models within the region below $\theta^l$, while utilizing a new reset mechanism called Precise-Positioning Reset (P2-Reset) in the region above $\theta^l$. Specifically, after each firing process, HD-LIF model will reset the value of $\mathbf{v}^l$ to $\theta^l$ and transmit spikes corresponding to the reset strength:
\begin{align}
    \mathbf{m}_t^l &= \lambda_t^l \odot \mathbf{v}_{(t-1)}^l + \mathbf{I}_{t}^l, \
    \mathbf{v}_{t}^l = \mathbf{m}_t^l - \mathbf{s}_{t}^{l,\ast}, \nonumber \\
    \mathbf{I}_{t}^l &= \hat{\mathbf{W}}^l\mathbf{s}_{t}^{(l-1)}, \
    \mathbf{s}_{t}^{l,\ast} = \left\{
        \begin{aligned}
        &\mathbf{m}_t^l \!-\! \theta_t^l,\ \mathbf{m}_t^l \geq \theta_t^l \\
        &0,\ \text{otherwise}
        \end{aligned}
    \right., \nonumber \\
    \mathbf{s}_t^l &= \mathbf{Q}(\mathbf{s}_{t}^{l,\ast},s,n,\tau) = s\!\cdot\! \text{clip}\!\left( \left\lfloor \frac{\mathbf{s}_{t}^{l,\ast}}{s} \right\rceil_{\tau}, 0, 2^n\!-\!1\!+\!\tau \right)\!.
    \label{eq01}
\end{align}
Here $\hat{\mathbf{W}}^l$ adopts the learning modes of 1-bit \citep{liu2021adam} or 1.5-bit \citep{li2016ternary}, which compresses the original weight to a ultra-low number of bits. Specifically, the 1-bit scheme ($\{-1, +1\}$) aims to obtain the optimal compression for synaptic parameters, while the 1.5-bit scheme ($\{0, \pm1\}$) will further reduce the number of Synaptic Operations (SOPs) and power consumption by promoting the sparsity in synaptic weights. For $\lfloor x\rceil_\tau$, we define $\lfloor x\rceil_\tau=\lfloor x\rceil,\tau=0;\lfloor x\rceil_\tau=x,\tau\rightarrow+\infty$. $n$ and $s$ denote the compressed bit-width and scaling factor.
According to Eq.(\ref{eq01}), we can further derive the back-propagation chain of HD-LIF model during the online training process:
\begin{align}
    \nabla_{\mathbf{W}^l} \mathcal{L} \!= \!\sum_{t=1}^T \frac{\partial \mathcal{L}}{\partial \mathbf{m}_t^l}\!\frac{\partial \mathbf{m}_t^l}{\partial \hat{\mathbf{W}}^l}\!\frac{\partial \hat{\mathbf{W}}^l}{\partial \mathbf{W}^l},\ 
    \frac{\partial \mathcal{L}}{\partial \mathbf{m}_t^l} \!=\! \frac{\partial \mathcal{L}}{\partial \mathbf{s}_{t}^l}\!\frac{\partial \mathbf{s}_{t}^l}{\partial \mathbf{s}_t^{l,\ast}}\!\frac{\partial \mathbf{s}_t^{l,\ast}}{\partial \mathbf{m}_t^l}.    
    \label{eq01-b}
\end{align}
For HD-LIF model, on the one hand, $\frac{\partial \mathbf{s}_t^{l,\ast}}{\partial \mathbf{m}_t^l}$ remains constant in $(-\infty, \theta_t^l]$ and $(\theta_t^l, +\infty)$ respectively, which makes the surrogate gradient of HD-LIF model independent of the corresponding membrane potential value, improving gradient separability along the temporal dimension; on the other hand, there is no non-differentiable problem between $\mathbf{s}_t^l$ and $\mathbf{m}_t^l$ during the firing process, ensuring gradient alignment along the spatial dimension.
Based on the above observations, we can further propose the following theorem: 
\begin{theorem}
\label{thm01}
    For HD-LIF model under the condition of online training, combining with Definition 4.1, we will have: $\forall t,i\in[1,T], t<i, \bm{\epsilon}^l[i,t] = \bm{\chi}^l[i,i] \prod_{j=t+1}^i \bm{\chi}^l[j,j\!-\!1]$, here we define $\bm{\chi}^l[i,i]=\frac{\partial \mathbf{s}_i^l}{\partial \mathbf{m}_i^l} \in\left\{ 0,1 \right\}, \bm{\chi}^l[j,j\!-\!1]=\frac{\partial \mathbf{m}_j^l}{\partial \mathbf{m}_{j-1}^l} \in\left\{ 0,\lambda_j^l \right\}$.\\
    (\romannumeral1) If $\exists k\in[t,T], \bm{\chi}^l[k,k]=1$, then we can directly derive the gradient relationship between STBP and online training: $\left( \frac{\partial \mathcal{L}}{\partial \mathbf{m}_t^l} \right)_{\textit{Online}} = \frac{\bm{\chi}^l[t,t]}{\left( \bm{\chi}^l[t,t] + \sum_{i=t+1}^T \bm{\chi}^l[i,i] \prod_{j=t+1}^i \bm{\chi}^l[j,j\!-\!1] \right)} \left( \frac{\partial \mathcal{L}}{\partial \mathbf{m}_t^l} \right)_{\textit{STBP}}$.\\
    (\romannumeral2) Under more general conditions, $\left( \frac{\partial \mathcal{L}}{\partial \mathbf{m}_t^l} \right)_{\textit{STBP}} \in \left\{ 1, \prod_{j=t+1}^{t^{\ast}} \lambda_j^l\right\} \cdot \left( \frac{\partial \mathcal{L}}{\partial \mathbf{s}_t^l} \right)$, here $\bm{\chi}^l[t^{\ast},t^{\ast}]=1\ \land \ \forall k\in[t, t^{\ast}),\bm{\chi}^l[k,k]=0$.
\end{theorem}
From Theorem \ref{thm01}, one can find that HD-LIF model can convert $\bm{\epsilon}^l[i,t]$ into a constant value belonging to a finite set, thereby achieving the separability of the backward gradients. On this basis, we set $\lambda_t^l$ and $\theta_t^l$ as learnable membrane-related parameters for HD-LIF model at each time-step, enabling HD-LIF model to more adaptively regulate its learning gradients during online training.

\subsection{HD-LIF model family}
Eq.(\ref{eq01}) describes the dynamic process of vanilla HD-LIF model, consisting of the regions shown in Fig.\ref{fig02}(\uppercase\expandafter{\romannumeral1})-(\uppercase\expandafter{\romannumeral3}). Although vanilla model is already enable to engage in online learning, we continue to explore more possible dynamic calculation schemes within the model based on this foundation, aiming to further enhance the computational efficiency and learning capability of HD-LIF model. We will combine the following variant versions of HD-LIF model with the vanilla version to form the HD-LIF model family, as illustrated in Fig.\ref{fig03}.

\textbf{Parallel HD-LIF model.} Due to the fact that vanilla HD-LIF model is based on traditional serial computing processes, there will still be $T$ MUL (for leakage process) and $2T$ ADD (for charging and resetting process) operations within the model during the inference stage. In other words, compared to the LIF model, vanilla HD-LIF model has no advantage in the inference overhead of neuron layers. Therefore, as shown in Fig.\ref{fig03}(a)-(b), we can consider introducing a parallel version of the HD-LIF model at an appropriate ratio, \ie, setting $\mathbf{s}_t^{l,\ast}:=(\mathbf{I}_t^l\geq\theta_t^l)$. At this point, there is no leakage or charging process for neuron layers, NOPs are only composed of $T$ ADD operations.

\textbf{Mem-BN HD-LIF model.} Conventional BN layers in SNNs are usually placed behind the corresponding synaptic convolutional layers along the spatial dimension to monitor the distribution of input current $[\mathbf{I}_1^l,...,\mathbf{I}_T^l]$, ensuring the stability of STBP training and seamlessly integrating with the convolutional layers during the inference stage. However, for online training, considering the lack of temporal gradient terms, in addition to controlling the distribution of input current, we also need to pay attention to the distribution stability of the cumulative membrane potential $[\mathbf{m}_1^l,...,\mathbf{m}_T^l]$ over time-steps. Thanks to the fusible property of the BN layer, we consider a batch-normalization scheme oriented towards membrane potential along the temporal dimension, which can exquisitely fuse BN layers with membrane-related parameters in the SNN inference stage, further enhancing the learning ability of HD-LIF model without introducing additional computational costs, as shown in Fig.\ref{fig02}(\uppercase\expandafter{\romannumeral4}). Specifically, we introduce the following calculation process for Mem-BN in addition to Eq.(\ref{eq01}):
\begin{align}
    &\hat{\mathbf{m}}_t^l = \alpha_t^l \odot \mathbf{m}_t^l + \beta_t^l \odot \textbf{BN}_t^l(\mathbf{m}_t^l),\ \mathbf{v}_t^l = \hat{\mathbf{m}}_t^l - \mathbf{s}_t^{l,\ast}, \nonumber \\
    &\textbf{BN}_t^l(\mathbf{m}_t^l) = \gamma^l \cdot \frac{\mathbf{m}_t^l - \mu^l_t}{\sqrt{{\sigma^l_t}^2 + \epsilon}} + b^l.
    \label{eq02}
\end{align}
Here $\mu_t^l$ and $\sigma_t^l$ are the recorded mean and standard deviation of $\mathbf{m}_t^l$, while $\gamma_t^l$ and $\epsilon, b_t^l$ denote the scaling and shifting factors of the BN layer. $\alpha_t^l$ and $\beta_t^l$ are learnable parameters that regulate the normalization degree of $\mathbf{m}_t^l$. When $\alpha_t^l=1, \beta_t^l=0$, the Mem-BN version will degrade to the vanilla version mentioned in \S \ref{section01}, which ensures its performance lower-bound. 

For SNN inference, Mem-BN HD-LIF can be equivalently converted into vanilla HD-LIF model through the following re-parameterization process. Specifically, Eq.(\ref{eq02}) can be first integrated into the following equation:
\begin{align}
    \hat{\mathbf{m}}_t^l = (\alpha_t^l \!+\! \frac{\gamma_t^l \cdot \beta_t^l}{\sqrt{{\sigma_t^l}^2 + \epsilon}}) \mathbf{m}_t^l - \beta_t^l (\frac{\gamma_t^l \cdot \mu_t^l}{\sqrt{{\sigma_t^l}^2 + \epsilon}} \!-\! b_t^l).
    \label{eq03}
\end{align}
Subsequently, we can merge the scaling and shifting terms \wrt $\mathbf{m}_t^l$ in Eq.(\ref{eq03}) into membrane-related parameters at different positions, including membrane leakage parameters $[\lambda_1^l,...,\lambda_T^l]$ and input current $[\mathbf{I}_1^l,...,\mathbf{I}_t^l]$, so that Mem-BN HD-LIF still only involves the processes of leakage, charging and resetting during inference calculations. Here we set $\alpha_t^{l,\ast}=\alpha_t^l + \frac{\gamma_t^l \cdot \beta_t^l}{\sqrt{{\sigma_t^l}^2 + \epsilon}}, \beta_t^{l,\ast}=\beta_t^l (\frac{\gamma_t^l \cdot \mu_t^l}{\sqrt{{\sigma_t^l}^2 + \epsilon}} - b_t^l)$ and derive the final membrane-related parameters $\forall t, \hat{\lambda}_t^l=\alpha_t^{l,\ast} \lambda_t^l, \hat{\mathbf{I}}_t^l=\alpha_t^{l,\ast}\mathbf{I}_t^l-\beta_t^{l,\ast}$ after re-parameterization:
\begin{align}
    \left( \hat{\mathbf{m}}_t^l \right)_\text{inference} &= \alpha_t^{l,\ast} \mathbf{m}_t^l - \beta_t^{l,\ast} \nonumber \\
    &= \alpha_t^{l,\ast} (\lambda_t^l \odot \mathbf{v}_{(t-1)}^l + \mathbf{I}_t^l) - \beta_t^{l,\ast} \nonumber \\
    &= (\alpha_t^{l,\ast} \lambda_t^l) \odot \mathbf{v}_{(t-1)}^l + (\alpha_t^{l,\ast}\mathbf{I}_t^l-\beta_t^{l,\ast}) \nonumber \\
    &= \hat{\lambda}_t^l \odot \mathbf{v}_{(t-1)}^l + \hat{\mathbf{I}}_t^l
    \label{eq04}
\end{align}

\subsection{Enhancing HD-LIF through efficient attention mechanism}
\begin{table*}[t]
    \caption{Comparison with previous SoTA works from SNN mainstream or online learning paradigms. Here Param.(MB) denotes the parameter memory of convolutional and linear layers in the corresponding standard network backbone.}
    \renewcommand\arraystretch{0.95}
	\centering
    \resizebox{0.9\linewidth}{!}{
    \begin{threeparttable}
	   \begin{tabular}{c|cccccc} \Xhline{1pt}
        \textbf{Dataset} & \textbf{Method} & \textbf{Architecture} & \textbf{Param.(MB)} & \textbf{Type} &  \textbf{Time-steps} & \textbf{Accuracy(\%)} \\ \hline
        \multirow{6}{*}{CIFAR-10} & STBP-tdBN \cite{zheng2021going} & ResNet-19 & 50.48 & STBP Training & 4 & 92.92 \\
        & Dspike \cite{li2021dspike} & ResNet-18 & 44.66 & STBP Training & 4 & 93.66 \\
        & TET \cite{deng2022temporal} & ResNet-19 & 50.48 & STBP Training & 4 & 94.44 \\
        & GLIF \cite{yao2022GLIF} & ResNet-18 & 44.66 & STBP Training & 4, 6 & 94.67, 94.88 \\
        & SLTT \cite{Meng2023SLTT} & ResNet-18 & 44.66 & Online Training & 6 & 94.44 \\
        & \cellcolor{Periwinkle!5} \textcolor{Periwinkle}{\textbf{Ours}} & \cellcolor{Periwinkle!5} \textcolor{Periwinkle}{\textbf{ResNet-18}} & \cellcolor{Periwinkle!5} \textcolor{Periwinkle}{\textbf{2.82}} & \cellcolor{Periwinkle!5} \textcolor{Periwinkle}{\textbf{Online Training}} & \cellcolor{Periwinkle!5} \textcolor{Periwinkle}{\textbf{4}} & \cellcolor{Periwinkle!5} \textcolor{Periwinkle}{\textbf{95.59}} \\ \hline
        
        \multirow{5}{*}{CIFAR-100} & Dspike \cite{li2021dspike} & ResNet-18 & 44.84 & STBP Training & 4 & 73.35 \\
        & TET \cite{deng2022temporal} & ResNet-19 & 50.57 & STBP Training & 4 & 74.47 \\
        & GLIF \cite{yao2022GLIF} & ResNet-18 & 44.84 & STBP Training & 4, 6 & 76.42, 77.28 \\
        & SLTT \cite{Meng2023SLTT} & ResNet-18 & 44.84 & Online Training & 6 & 74.38 \\
        & \cellcolor{Periwinkle!5} \textcolor{Periwinkle}{\textbf{Ours}} & \cellcolor{Periwinkle!5} \textcolor{Periwinkle}{\textbf{ResNet-18}} & \cellcolor{Periwinkle!5} \textcolor{Periwinkle}{\textbf{3.00}} & \cellcolor{Periwinkle!5} \textcolor{Periwinkle}{\textbf{Online Training}} & \cellcolor{Periwinkle!5} \textcolor{Periwinkle}{\textbf{4}} & \cellcolor{Periwinkle!5} \textcolor{Periwinkle}{\textbf{78.45}} \\ \hline

        \multirow{5}{*}{ImageNet-200} & DCT \cite{garg2021dct} & VGG-13\tnote{$\dag$} & 38.02$^{\ast}$ & Hybrid Training & 125 & 56.90 \\
        & Offline-LTL \cite{yang2022tandem} & \multirow{2}{*}{VGG-13\tnote{$\dag$}} & \multirow{2}{*}{38.02$^{\ast}$} & STBP Training & 16 & 55.37 \\
        & Online-LTL \cite{yang2022tandem} & & & Online Training & 16 & 54.82 \\
        & ASGL \cite{wang2023ASGL} & VGG-13 & 38.02 & STBP Training & 4, 8 & 56.57, 56.81 \\
        & \cellcolor{Periwinkle!5} \textcolor{Periwinkle}{\textbf{Ours}} & \cellcolor{Periwinkle!5} \textcolor{Periwinkle}{\textbf{VGG-13}} & \cellcolor{Periwinkle!5} \textcolor{Periwinkle}{\textbf{2.77}} & \cellcolor{Periwinkle!5} \textcolor{Periwinkle}{\textbf{Online Training}} & \cellcolor{Periwinkle!5} \textcolor{Periwinkle}{\textbf{4}} & \cellcolor{Periwinkle!5} \textcolor{Periwinkle}{\textbf{60.68}} \\ \hline
        
        \multirow{5}{*}{ImageNet-1k} & STBP-tdBN \cite{zheng2021going} & ResNet-34 & 87.12 & STBP Training & 6 & 63.72 \\
        & TET \cite{deng2022temporal} & ResNet-34 & 87.12 & STBP Training & 6 & 64.79 \\
        & OTTT \cite{xiao2022OTTT} & ResNet-34 & 87.12 & Online Training & 6 & 65.15 \\
        & SLTT \cite{Meng2023SLTT} & ResNet-34 & 87.12 & Online Training & 6 & 66.19 \\
        & \cellcolor{Periwinkle!5} \textcolor{Periwinkle}{\textbf{Ours}} & \cellcolor{Periwinkle!5} \textcolor{Periwinkle}{\textbf{ResNet-34}} & \cellcolor{Periwinkle!5} \textcolor{Periwinkle}{\textbf{10.06}} & \cellcolor{Periwinkle!5} \textcolor{Periwinkle}{\textbf{Online Training}} & \cellcolor{Periwinkle!5} \textcolor{Periwinkle}{\textbf{4}} & \cellcolor{Periwinkle!5} \textcolor{Periwinkle}{\textbf{69.77}} \\ \hline
        
        \multirow{6}{*}{DVS-CIFAR10} & STBP-tdBN \cite{zheng2021going} & ResNet-19 & 50.48 & STBP Training & 10 & 67.80 \\
        & Dspike \cite{li2021dspike} & ResNet-18 & 44.66 & STBP Training & 10 & 75.40 \\
        & OTTT \cite{xiao2022OTTT} & VGG-SNN & 37.05 & Online Training & 10 & 76.30 \\
        & NDOT \cite{jiang2024ndot} & VGG-SNN & 37.05 & Online Training & 10 & 77.50 \\
        & \cellcolor{Periwinkle!5} & \cellcolor{Periwinkle!5} \textcolor{Periwinkle}{\textbf{ResNet-18}} & \cellcolor{Periwinkle!5} \textcolor{Periwinkle}{\textbf{2.81}} & \cellcolor{Periwinkle!5} & \cellcolor{Periwinkle!5} \textcolor{Periwinkle}{\textbf{10}} & \cellcolor{Periwinkle!5} \textcolor{Periwinkle}{\textbf{81.70}} \\
        & \cellcolor{Periwinkle!5} \multirow{-2}{*}{\textcolor{Periwinkle}{\textbf{Ours}}} & \cellcolor{Periwinkle!5} \textcolor{Periwinkle}{\textbf{VGG-SNN}} & \cellcolor{Periwinkle!5} \textcolor{Periwinkle}{\textbf{2.49}} & \cellcolor{Periwinkle!5} \multirow{-2}{*}{\textcolor{Periwinkle}{\textbf{Online Training}}} & \cellcolor{Periwinkle!5} \textcolor{Periwinkle}{\textbf{10}} & \cellcolor{Periwinkle!5} \textcolor{Periwinkle}{\textbf{83.00}} \\ \Xhline{1pt}
	\end{tabular}
    \begin{tablenotes}
    \small
    \item[$\dag$] utilizes fully-connected layers with relatively large parameter quantity.
    \end{tablenotes}
    \end{threeparttable}
    }
	\label{table01}
\end{table*}
Channel Attention mechanism \citep{hu2018squeeze, wang2020eca, guo2022join} is usually inserted after the convolutional layers to further optimize the network performance. Considering that it is not suitable to introduce computation- or parameter-intensive attention mechanisms during online training and deployment, we transfer the idea of ECA \citep{wang2020eca} to the HD-LIF online training framework, then propose Spiking Efficient Channel Attention (SECA) mechanism, as shown in Eq.(\ref{eq05}). 
\begin{align}
    \text{SECA}(\mathbf{I}_t^l) &= \sigma\ (\ \underbrace{\text{Conv1d}}_{\in\ \mathbb{R}^{1\times 1\times K}}\ (\ \underbrace{\text{GAP}\ (\ \mathbf{I}_t^l\ )}_{\in\ \mathbb{R}^{B\times 1\times C}}\ )\ ) \odot \mathbf{I}_t^l.
    \label{eq05}
\end{align}
Here the input current $\mathbf{I}_t^l\in \mathbb{R}^{B\times C\times H\times W}$ will be compressed to $\mathbb{R}^{B\times C\times 1\times 1}$ through the Global Average Pooling (GAP) layer, then $\text{Conv1d}(\cdot)$ and Sigmoid Function $\sigma(\cdot)$ will be used to capture and activate the attention scores among different channels, ultimately merging with the shortcut path.
Considering that the HD-LIF model can convey enough information representation at each time-step, we enable the spike sequence to share the weight of SECA in the temporal dimension. Due to its ultra-low parameter quantity and computational load (1 Conv1d layer with $O(K)$ parameters and $O(KC)$ operations), SECA can further enhance the learning ability of HD-LIF SNNs under the condition of hardly affecting its online training and deployment.
In addition, as shown in Fig.\ref{fig03}(c)-(d) and Eq.(\ref{eq06}), we further propose two variants for SECA:
\begin{align}
    \text{SECA}_\text{\uppercase\expandafter{\romannumeral1}}^l(\cdot):\text{SECA}(\mathbf{I}_t^l),\ \text{SECA}_\text{\uppercase\expandafter{\romannumeral2}}^l(\cdot):\text{SECA}(\mathbf{I}_t^{(l-1)} \!+\! \mathbf{I}_t^l).
    \label{eq06}
\end{align}
Here $\text{SECA}_\text{\uppercase\expandafter{\romannumeral1}}^l(\cdot)$ is the vanilla channel attention mechanism, while considering the shortcomings of compressed synaptic layers in feature extraction, $\text{SECA}_\text{\uppercase\expandafter{\romannumeral2}}^l(\cdot)$ combines the input currents from both pre-synaptic and post-synaptic layers to further enhance the effectiveness of the attention mechanism.
To achieve efficient online training for HD-LIF models, we draw inspiration from the scheme proposed in \cite{Meng2023SLTT}, where for each training batch, we merely perform gradient update for synaptic weights and corresponding membrane-related parameters at a randomly selected time-step. The overall procedure has been summarized in Algorithm \ref{alg01}.
\begin{algorithm}[t]
    \caption{Online learning framework for HD-LIF model family.}
    \label{alg01}
    \begin{algorithmic}[1]
    \REQUIRE SNN model $f_\text{SNN}(\mathbf{W}, \lambda, \theta, \alpha, \beta)$ with $L$ layers; training dataset $\mathbf{D}$; time-steps $T$
    \ENSURE Deployable SNN model $f_\text{SNN}(\hat{\mathbf{W}}, \hat{\lambda}, \theta)$
    \STATE \textbf{\textcolor{Periwinkle}{\# Online Training Stage}}
    \STATE Define gradient update time for training batches: $\mathbf{G}=\{ t\in\text{randint}_{1:T} \ \text{for}\ x_{data}\in\mathbf{D} \}$
    \FOR{$x_{data},x_{label}\in\mathbf{D}$; $t=1$ to $T$; $l=1$ to $L$}
        \STATE perform forward propagation $\in\{\mathbf{s}_t^l = \mathbf{Q}(\hat{\mathbf{W}}^l\mathbf{s}_t^{(l-1)} \!-\! \theta_t^l,s,n,\tau)$, Eq.(\ref{eq01}), Eq.(\ref{eq02})$\}$
        \STATE for SECA, calculate and merge the attention scores in Eqs.(\ref{eq05}-\ref{eq06})
        \IF{$t$ is equal to $\mathbf{G}_{x_{data}}$}
            \STATE perform back-propagation in Eq.(\ref{eq01-b}), update $\hat{\mathbf{W}}^l$ and learnable membrane-related parameters
        \ENDIF
    \ENDFOR
    \STATE \textbf{\textcolor{Periwinkle}{\# Online Deployment Stage}}
    \FOR{$t=1$ to $T$; $l=1$ to $L$}
        \STATE convert $\mathbf{W}^l$ to $\hat{\mathbf{W}}^l$
        \IF{current neuron layer is Mem-BN HD-LIF}
            \STATE re-parameterize $\lambda_t^l, \alpha_t^l, \beta_t^l, \textbf{BN}_t^l(\cdot)$ into $\hat{\lambda}_t^l, \hat{\mathbf{I}}_t^l$ through Eq.(\ref{eq04})
        \ENDIF
    \ENDFOR
    \end{algorithmic}
\end{algorithm}
\section{Experiments}
To validate the superiority of the HD-LIF online learning framework, we evaluate its performance across a diverse set of datasets spanning different scales and types, including CIFAR-10/100 \citep{Krizhevsky2009CIFAR100}, ImageNet-200/1K \citep{Deng2009ImageNet} and DVS-CIFAR10 \citep{li2017cifar10}. We compare against representative prior works in the domain of SNN mainstream and online training on ResNet \citep{he2016deep, hu2021residual} and VGG \citep{Simonyan2014VGG16} backbones.
Specifically, all ResNet cases on static datasets use the block structure of Fig.\ref{fig03}(b) by default; when adopting neuromorphic data and SECA, we replace vanilla HD-LIF in the block with Mem-BN HD-LIF to further enhance the learning ability of SNNs. The SECA module based on ResNet chooses the structure mentioned in Fig.\ref{fig03}(d). For VGG structure, we utilize the vanilla HD-LIF model for static datasets and Mem-BN version for neuromorphic datasets.
More training and implementation details have been provided in Appendix.

\subsection{Comparison with previous SoTA works}
\begin{table*}[t]
    \caption{Ablation studies for HD-LIF models with different configuration on CIFAR-100, ResNet-18.}
    \renewcommand\arraystretch{1.0}
	\centering
    \resizebox{0.9\linewidth}{!}{
    \begin{threeparttable}
	\begin{tabular}{c|cc|ccccc} \Xhline{0.5pt}
        \textbf{Model Config.} & \textbf{T. Speed(s/epoch)} & \textbf{GPU Mem.(GB)} & \textbf{Param.(MB)} & \textbf{Acc.(\%)} & \textbf{SOPs(M)} & \textbf{NOPs(M)} & \textbf{Power(mJ)} \\ \hline
        \faint{LIF} & \faint{20.52} & \faint{1.50} & \faint{44.84} & \faint{71.75} & \faint{273.02} & \faint{6.59} & \faint{0.25} \\
        HD-LIF & 37.93 & 1.68 & 4.40 & \textbf{80.16} & 284.49 & 6.59 & 0.26 \\
        HD-LIF\tnote{$\dag$} & 40.91 & 1.92 & 4.40 & 79.62 & 233.84 & 6.59 & 0.03 \\
        HD-LIF\tnote{$\ddag$} & \textbf{35.39} & \textbf{1.44} & 4.40 & 78.82 & 254.08 & 4.62 & 0.23 \\
        HD-LIF\tnote{$\dag, \ddag$} & 38.50 & 1.70 & \textbf{4.40} & 78.61 & \textbf{190.13} & \textbf{4.62} & \textbf{0.02} \\
        \Xhline{0.5pt}
	\end{tabular}
    \begin{tablenotes}
    \small
    \item[$\dag$] denotes compressing the firing spikes to 4-bits, \ie $\tau=0,n=4$.
    \item[$\ddag$] denotes substituting vanilla HD-LIF with Parallel HD-LIF at a proportion of 50\%.
    \end{tablenotes}
    \end{threeparttable}
    }
	\label{table02}
\end{table*}
\begin{table*}[h]
    \caption{Parameter memory and accuracy of SNNs before and after utilizing SECA.}
    \renewcommand\arraystretch{1.0}
	\centering
        \resizebox{0.9\linewidth}{!}{
	\begin{tabular}{c|cc|cc|cc} \Xhline{0.5pt}
        \multirow{2}{*}{Method} & \multicolumn{2}{c|}{CIFAR-10, ResNet-18} & \multicolumn{2}{c|}{CIFAR-100, ResNet-18} & \multicolumn{2}{c}{DVS-CIFAR10, ResNet-18} \\ \cline{2-7}
        & Param.(MB) & Acc.(\%) & Param.(MB) & Acc.(\%) & Param.(MB) & Acc.(\%) \\ \hline
        \faint{HD-LIF} & \faint{2.82} & \faint{95.59} & \faint{3.00} & \faint{78.45} & \faint{2.81} & \faint{81.70} \\
        \textbf{HD-LIF (Mem-BN \& SECA)} & \textbf{2.82} & \textbf{95.91 \textcolor{Periwinkle}{(+0.32)}} & \textbf{3.00} & \textbf{79.33 \textcolor{Periwinkle}{(+0.88)}} & \textbf{2.81} & \textbf{83.50 \textcolor{Periwinkle}{(+1.80)}} \\
        \Xhline{0.5pt}
	\end{tabular}}
	\label{table03}
\end{table*}
\textbf{CIFAR-10 \& CIFAR-100.} As shown in Tab.\ref{table01}, compared to conventional STBP and online learning framework, our scheme utilizes advanced spiking calculation mechanism and synaptic weight compression technique, which saves approximately $15\times$ synaptic parameter memory without performance degradation. In addition, we achieve higher learning precision within the same or fewer time-steps. For example, our method outperforms GLIF \citep{yao2022GLIF} and SLTT \citep{Meng2023SLTT} with accuracies of 2.03\% and 4.07\% respectively on CIFAR-100, ResNet-18.

\textbf{ImageNet-200 \& ImageNet-1k.} For large-scale datasets, our HD-LIF model has also demonstrated significant advantages. For instance, we respectively achieve accuracies of 60.68\% and 69.77\% on ImageNet-200 and ImageNet-1k, which exceeds the corresponding online learning methods \citep{yang2022tandem, Meng2023SLTT} within fewer time-steps and saves more than 88\% of synaptic parameter memory.

\textbf{DVS-CIFAR10.} Our method can also achieve effective information extraction for neuromorphic data. From Tab.\ref{table02}, one can note that our learning precision is 6.30\% higher than Dspike \citep{li2021dspike} and 5.50\% higher than NDOT \citep{jiang2024ndot} under the condition of utilizing the identical network backbone and time-steps.

\subsection{Detailed analysis for performance metric}
Table \ref{table02} shows the comparison of online learning frameworks based on LIF and HD-LIF model in multiple performance metrics of training and inference stages. For Speed and GPU memory during the training stage, since HD-LIF involves more learnable parameters, our approach incurs slightly higher overhead than vanilla LIF model, but the total computational overhead still remains at a constant-level complexity. In the inference stage, the HD-LIF framework can significantly surpass vanilla online training in parameter memory and inference accuracy, combining with parallel calculation mode, it can further save 30\% NOPs. In addition, it is worth noting that HD-LIF can maintain superior learning ability even when significantly compressing the information density of firing spikes, thereby achieving low-memory, low-power and high-precision online inference. For example, HD-LIF can outperform LIF by 7.87\% while saving 10$\times$ parameter memory and 9$\times$ power consumption.

\subsection{Performance validation for SECA}
As shown in Tab.\ref{table03}, we explore the network performance before and after inserting SECA modules for multiple datasets. One can note that SECA hardly introduces additional parameter quantity, but can provide extra precision improvement for synaptic layers. Among them, after introducing SECA mechanism, the performance improvement of HD-LIF online learning on DVS-CIFAR10, ResNet-18 even reaches 1.80\%, which effectively compensates for the potential accuracy loss caused by the compression of synaptic weights.

\subsection{Temporal processing property of HD-LIF}
\begin{figure} [t]\centering  
\includegraphics[width=0.96\columnwidth]{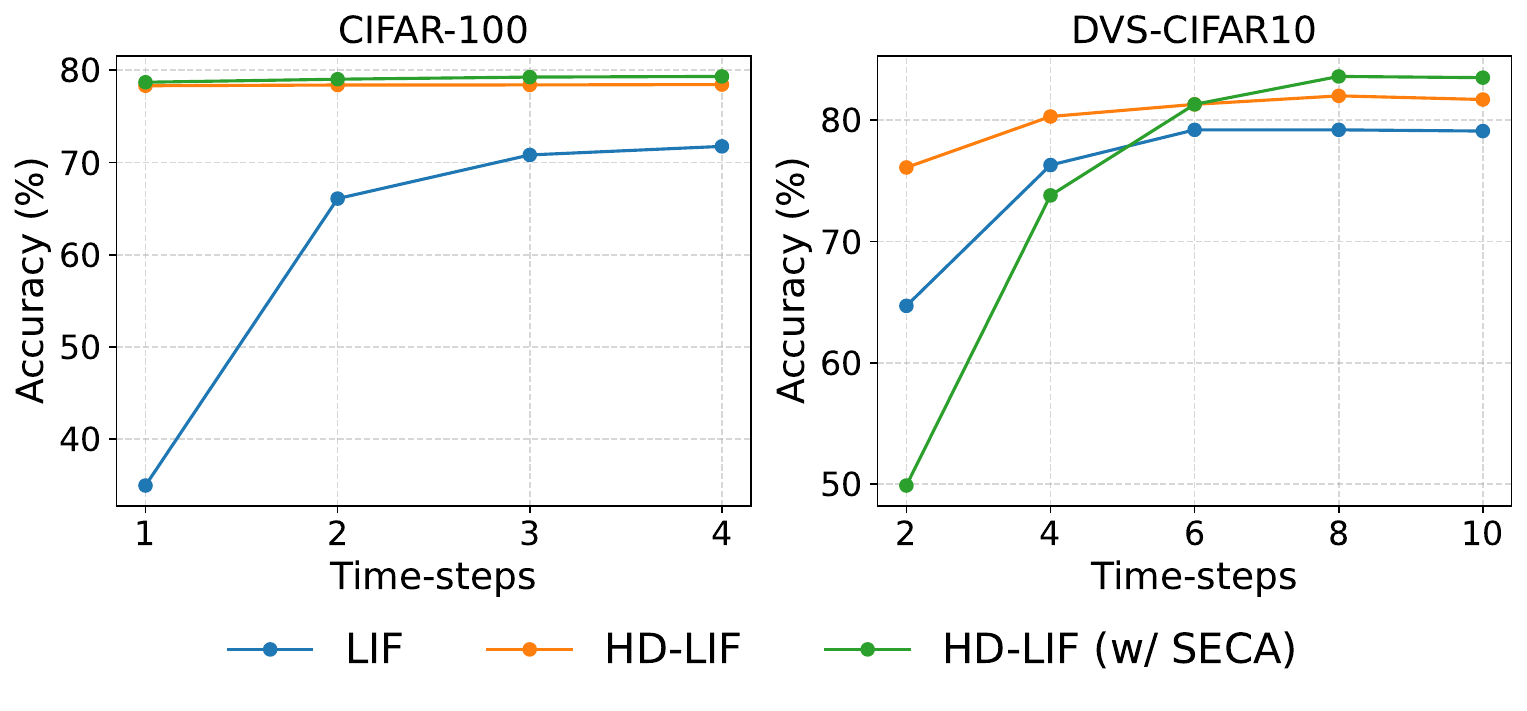}
\caption{Accuracies of HD-LIF \& LIF models over time-steps.}
\label{fig04}
\end{figure}
As shown in Fig.\ref{fig04}, due to the ability of HD-LIF to transmit richer spiking information at each time-step, for static datasets, HD-LIF models are enable to reach approximately SoTA performance at the first time-step, reflecting their similarity to ANN; for neuromorphic data, HD-LIF can gradually accumulate information along the temporal dimension, resulting in an increase in inference accuracy with the number of time-steps, which maintains the same property as vanilla spiking models. The above results demonstrate the duality of the hybrid-driven mechanism. Here Tab.\ref{table02} and Fig.\ref{fig04} adopt unified LIF baselines.
\section{Conclusions}
In this paper, we systematically analyze the deficiencies of conventional online training, then propose a novel online learning framework based on the HD-LIF model family, which effectively tackles the gradient inseparability and misalignment problem. By flexibly combining various optimization techniques, HD-LIF model can further achieve comprehensive online learning and deployment towards multiple performance evaluation metrics. Experimental results have verified that our proposed scheme can break through the limitations of the previous learning paradigm and provide further inspiration for the future development of online learning.

\section*{Acknowledgments}
This work is supported by the National Natural Science Foundation of China (U24B20140, 62422601, 62506011), Beijing Municipal Science and Technology Program (Z241100004224004), Beijing Nova Program (20230484362, 20240484703).

{
    \small
    \bibliographystyle{ieeenat_fullname}
    \bibliography{main}
}


\end{document}